\documentclass[a4paper,conference]{IEEEtran}
\ifCLASSINFOpdf
\usepackage[pdftex]{graphicx}
\DeclareGraphicsExtensions{.pdf,.jpeg,.png, .eps}
\else
\fi
\usepackage[labelformat=simple]{subcaption}

\usepackage{amsmath, amssymb}
\usepackage{mathrsfs}
\usepackage{xcolor}
\usepackage{array}
\usepackage{multirow}
\begin{document}
%
\title{VPTR: Efficient Transformers for Video Prediction}

\author{\IEEEauthorblockN{Xi Ye}
\IEEEauthorblockA{LITIV Laboratory, Polytechnique Montréal\\
Montréal, Canada\\
Email: xi.ye@polymtl.ca}
\and
\IEEEauthorblockN{Guillaume-Alexandre Bilodeau}
\IEEEauthorblockA{LITIV Laboratory, Polytechnique Montréal\\
Montréal, Canada\\
Email: gabilodeau@polymtl.ca}}


%


\maketitle

\begin{abstract}
In this paper, we propose a new Transformer block for video future frames prediction based on an efficient local spatial-temporal separation attention mechanism. Based on this new Transformer block, a fully autoregressive video future frames prediction Transformer is proposed. In addition, a non-autoregressive video prediction Transformer is also proposed to increase the inference speed and reduce the accumulated inference errors of its autoregressive counterpart. In order to avoid the prediction of very similar future frames, a contrastive feature loss is applied to maximize the mutual information between predicted and ground-truth future frame features. This work is the first that makes a formal comparison of the two types of attention-based video future frames prediction models over different scenarios. The proposed models reach a performance competitive with more complex state-of-the-art models. The source code is available at \emph{https://github.com/XiYe20/VPTR}.

\end{abstract}


%
\IEEEpeerreviewmaketitle

\section{Introduction}

Video future frames prediction (VFFP) is applied to many research areas, for instance, intelligent agents ~\cite{liu2018a, lu2019}, autonomous vehicles \cite{bolte2019}, model-based reinforcement learning \cite{leibfried2016}. More recently, it has drawn a lot of attention since it is naturally a good self-supervised learning task \cite{bengio2013, wang2015}.

In this paper, we focus on the most common video prediction task, i.e. predicting $N$ future frames given $L$ past frames, with $L$ and $N$ greater than 1. For training a deep learning VFFP model, we can formalize the task to be $\arg\max_\theta p({\hat{x}_{L+N}, ..., \hat{x}_{L+1}}|x_L, ..., x_1; \theta)$, where $\hat{x}_t$ and $x_t$ denote the predicted future frames and input past frames respectively, $\theta$ denotes the model parameters.

Even though many deep learning-based VFFP models have been proposed, some challenges still remain to be solved. Almost all the state-of-the-art (SOTA) VFFP models are based on ConvLSTMs, i.e. convolutional short-term memory networks, which are efficient and powerful. Nevertheless, they suffer from some inherent problems of recurrent neural networks (RNNs), such as slow training and inference speed, error accumulation during inference, gradient vanishing, and predicted frames quality degradation. Researchers keep improving the performance by developing more and more sophisticated ConvLSTM-based models. For instance, by integrating custom motion-aware units into ConvLSTM \cite{chang2021}, or building complex memory modules to store the motion context \cite{lee2021}. 

Inspired by the great success of Transformers in NLP, more and more researchers are starting to adapt Transformers for various computer vision tasks \cite{meinhardt2021, dosovitskiy2021, esser2021, Arnab_2021_ICCV}, including few recent works for VFFP \cite{liu2020e,yan2021a, wu2021b}. However, it is computational expensive to apply Transformer to high dimensional visual features. We still need further research about more efficient visual Transformers, especially for videos. \emph{Therefore, we propose a novel efficient Transformer block with smaller complexity, and we developed a new video prediction Transformer (VPTR) based on it.}

Among the Transformers-based VFFP models \cite{liu2020e,yan2021a, wu2021b} that we mentioned earlier, some of them are autoregressive models while some others are non-autoregressive models, and they are based on different attention mechanisms, e.g. a custom convolution multi-head attention (MHA) \cite{liu2020e} and standard dot-product MHA \cite{yan2021a, wu2021b}. There is no formal comparison of the two typical approaches (autoregressive vs non-autoregressive) to use Transformer-based VFFP models so far. Thus, we developed an fully autoregressive VPTR (VPTR-FAR) and a non-autoregressive VPTR (VPTR-NAR). The two VPTR variants share the same attention mechanism and same number of Transformer block layers, which guarantees a fair comparison between the two approaches.

Our main contributions are summarized as:

1) We proposed a new efficient Transformer block for spatio-temporal feature learning by combining spatial local attention and temporal attention in two steps. The new Transformer block successfully reduces the complexity of a standard Transformer block with respect to same input spatio-temporal feature size, specifically, from $\mathcal{O}((THW)^2)$ to $\mathcal{O}(\frac{H^2W^2}{P^2} + T^2)$.

2) Two VPTR models, VPTR-NAR and VPTR-FAR, were developed. We show that the proposed simple attention-based VPTRs are capable of reaching and outperforming more complex SOTA ConvLSTM-based VFFP models.

3) A formal comparison of two VPTR variants was conducted. The results show that VPTR-NAR has a faster inference speed and smaller accumulation of errors during inference, but it is more difficult to train. We solved the training problem of VPTR-NAR by employing a contrastive feature loss which maximizes the mutual information of predicted and ground-truth future frame features.

4) We found that given the same number of Transformer block layers, VPTR-FAR has a worse generalization performance due to the accumulated inference errors, which are introduced by the discrepancy between train and test behaviors. We also found that recurrent inference over pixel space introduces less accumulation errors than recurrent inference over latent space in the case of VPTR-FAR.

\section{Related Work}

Almost all the SOTA deep learning-based VFFP models are ConvLSTM-based autoencoders, where the encoder extracts the representations of past frames, and then the decoder generates future frame pixels based on those representations \cite{chaabane2020, jin2020, wang2020, chang2021, lee2021}. In general, the SOTA models rely on complex ConvLSTM models that integrates attention mechanism or memory augmented modules. For example, LMC-Memory model \cite{lee2021} stores the long-term motion context by a novel memory alignment learning, and the motion information is recalled during test to facilitate the long-term prediction. Zhang et al. \cite{chang2021} proposed a attention-based motion-aware unit to increase the temporal receptive field of RNNs. 

The ConvLSTM-based models are flexible and efficient, but recurrent prediction is slow. Therefore, standard CNNs or 3D-CNNs are also used as the backbones of VFFP to generate multiple future frames simultaneously \cite{mathieu2016, chen2017a, vondrick2016, wu2020a}. Besides, the future prediction is by nature multimodal \cite{oprea2020a}, i.e. stochastic. Some VFFP models aim to solve this problem based on VAEs, such as SV2P \cite{babaeizadeh2018}, SVG-LP \cite{denton2018}, improved conditional VRNNs \cite{castrejon2019}. Stochasticity learning is challenging and thus most VFFP models ignore it. A detail survey of VFFP models can be found in \cite{oprea2020a}.

Recently, Transformers were applied for VFFP. The ConvTransformer \cite{liu2020e} model follows the architecture of DETR \cite{meinhardt2021}. DETR follows a classical neural machine translation (NMT) Transformer architecture. It also inspired the development of our VPTR-NAR. Despite the similarities, our VPTR-NAR is different from ConvTransformer with respect to the fundamental attention mechanism. Specifically, ConvTransformer proposed a custom hybrid multi-head attention module based on convolution, but our VPTR-NAR uses the standard multi-head dot-product attention. Another more recent model (VideoGPT) \cite{yan2021a} takes a 3D-CNN as backbone to encode video clips into spatial-temporal features, which are then flattened to be a sequence to train a standard Transformer with the GPT manner \cite{radford2018, radford2019language}. VideoGPT shares a similar architecture and train/test behaviours as our VPTR-FAR. But VideoGPT performs the attention along the spatial and temporal dimensions jointly while our VPTR-FAR performs the attention along the spatial and temporal dimensions separately. More importantly, VideoGPT downsamples the time dimension of input videos by 3D-CNN and thus helps the temporal information modeling. In contrast, our VPTR models solely depend on attention for a full temporal information modeling, without downsampling. Another recent work NÜWA \cite{wu2021b} shares a similar idea to VideoGPT.

\textbf{Efficient visual Transformers.} To reduce the computation cost for visual Transformers, some models reduce the flattened sequence length by different methods. ViT and the successive works \cite{dosovitskiy2021, wang2021, Arnab_2021_ICCV} divided input features into local patches, either 2D or 3D, and then tokenize the local patch by concatenation or Pooling \cite{fan2021}. Some other models introduce sparse attention to reduce the complexity, e.g. restricting the attention over a local region \cite{zhang2021, liu2021d, YuanFHLZCW21}, or decomposing the global attention into a series of axial-attention \cite{huang2019a, wang2020c, Arnab_2021_ICCV}. HRFormer \cite{YuanFHLZCW21} is an example of local region attention-based Transformers, which is designed for image classification and dense prediction.

Specifically, a HRFormer block is composed of a local-window multi-head self attention layer and a depth-wise convolution feed-forward network. The input feature maps $Z\in \mathbb{R}^{H \times W \times C}$ are firstly evenly divided into $P$ non-overlapping local patches, each patch is $Z_p\in \mathbb{R}^{\frac{H}{P} \times \frac{W}{P} \times C}$. Then a multi-head self attention is performed for each patch. Finally, the depth-wise convolution is used to exchange information among different local patches.

\section{The proposed VPTR models}
\label{sec: methods}

\subsection{Overall framework of VPTR}
Our overall VPTR framework is illustrated in Fig. \ref{fig:VPTR overall framework}. A CNN encoder shared by all the past frames extracts the visual features of each frame. Then a VPTR is taken to predict the visual features of each future frame based on the past frame features. The detail architectures of two different VPTR variants are described in the following subsections. In order to make the model architecture simple and easier to train, there is no skip connections between encoder and decoder.

\subsection{Encoder and decoder}

\begin{figure}[h]
\centering

\includegraphics[clip, trim=6.3cm 14.5cm 5cm 13cm, width=0.85\linewidth]{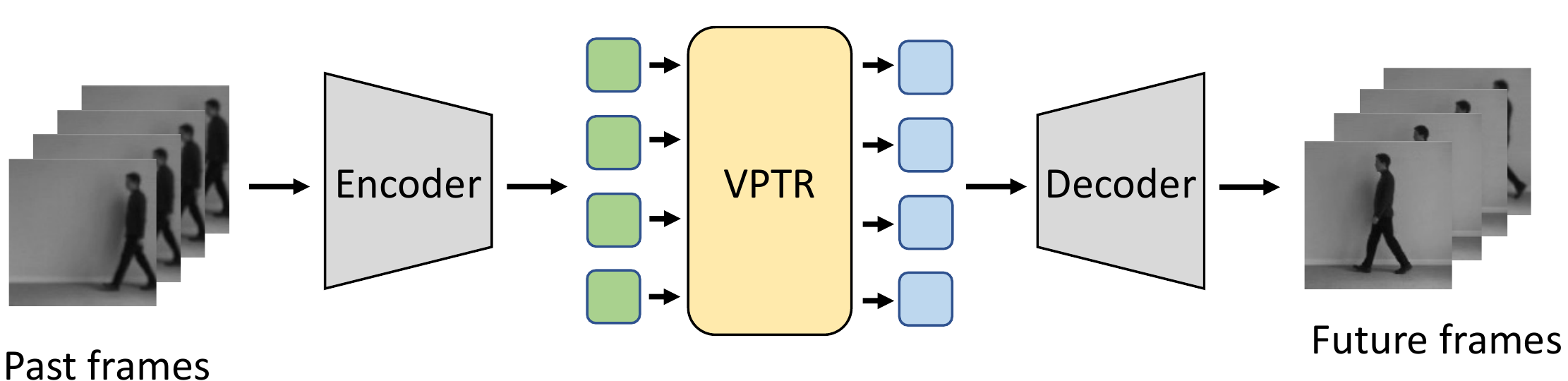}
\caption{Overall framework of VPTR. Green squares and blue squares denote the past frame features and future frames features respectively.}
\label{fig:VPTR overall framework}
\end{figure}

We adapted the ResNet-based autoencoder from the Pix2Pix model \cite{Isola2017}. The output feature channels of the encoder and input feature channels of the decoder are modified to be of size $d_{model}$ to match with the VPTR input and output size. The loss function to train the encoder and decoder is defined as follows,
\begin{multline}
    \mathcal{L}_{rec} = \mathcal{L}_2(X,\hat{X}) + \mathcal{L}_{gdl}(X, \hat{X})\\
    + \lambda_1 \arg\min_G\max_D  \mathcal{L}_{GAN}(G, D),
\label{eq: AE_loss}
\end{multline}
where $\mathcal{L}_2$ denotes the MSE loss (Eq. \ref{eq:l2}) and $\mathcal{L}_{gdl}$ denotes image gradient difference loss \cite{mathieu2016} (Eq. \ref{eq:gdl}), $X$ and $\hat{X}$ denote the original frames and reconstructed frames respectively, $x_i$ denotes a single frame, $\lambda_1$ and $\alpha$ are hyperparameters. $\mathcal{L}_{GAN}$ denotes the GAN loss (Eq. \ref{eq:gan_loss}), where $D$ denotes a discriminator, which is not shown in Fig. \ref{fig:VPTR overall framework}, and the combination of the encoder and decoder is considered to be a generator $G$. We train $\mathcal{L}_{GAN}$ with the PatchGAN \cite{Isola2017} manner. 

\begin{equation}
    \mathcal{L}_2(X,\hat{X}) = \sum_{i=1}^n \lVert x_i-\hat{x}_i \rVert _2^2
\label{eq:l2}
\end{equation}

\begin{multline}
    \mathcal{L}_{gdl}(X,\hat{X}) =\sum_{i=1}^n \sum_{i,j} \big | \lvert {x_{i,j}}-x_{i-1, j} \rvert - \lvert {\hat{x}_{i,j}}-\hat{x}_{i-1, j} \rvert \big |^\alpha\\
    + \big | \lvert {x_{i,j-1}}-x_{i, j} \rvert - \lvert {\hat{x}_{i,j-1}}-\hat{x}_{i, j} \rvert \big |^\alpha
\label{eq:gdl}
\end{multline}

\begin{equation}
    \mathcal{L}_{GAN}(G, D) = \mathbb{E}_X[logD(X)] + \mathbb{E}_{\hat{X}}[log(1 - D(G(X))]
\label{eq:gan_loss}
\end{equation}

\subsection{VidHRFormer Block}

\begin{figure*}[!h]
\centering
\subfloat[][]{
\includegraphics[clip, trim=11cm 7.35cm 11.5cm 7cm, width=0.1835\linewidth]{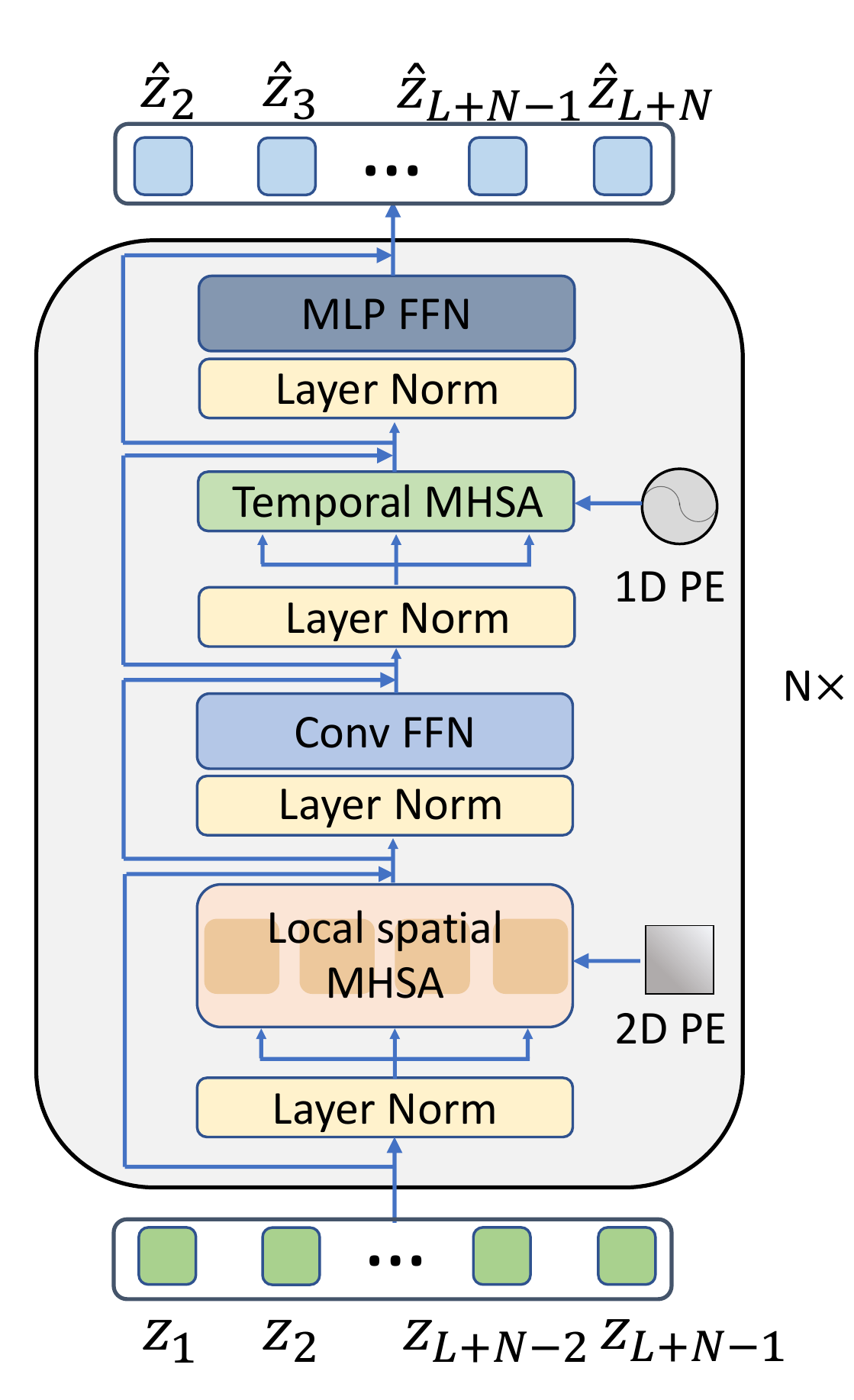}
\label{fig:FAR_model}}
\hfil
\subfloat[][]{
\includegraphics[clip, trim=3.5cm 3cm 7.7cm 5.2cm, width=0.366\linewidth]{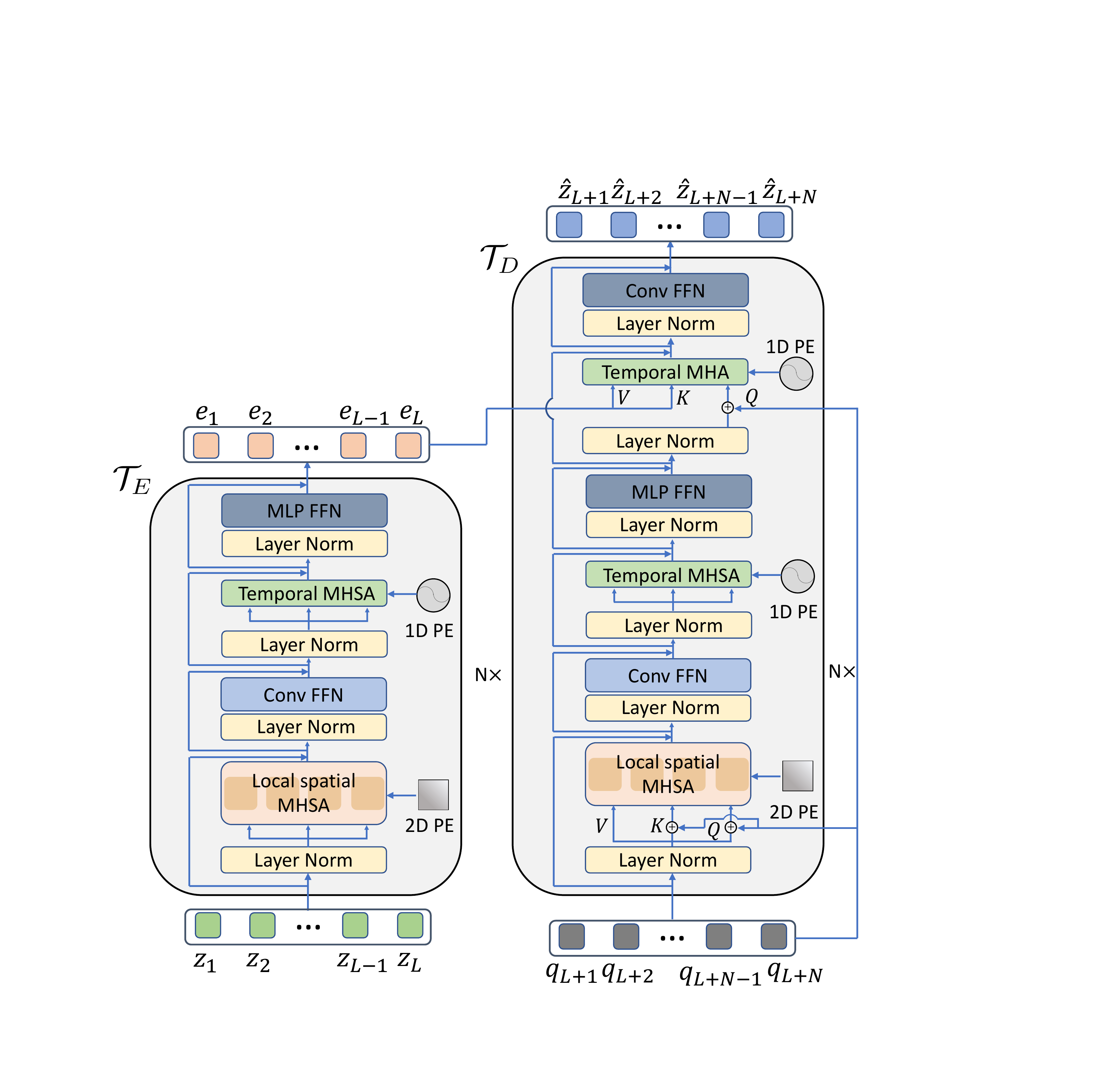}
\label{fig:NAR_model}}

\caption{(a) VPTR-FAR. The gray area indicates the proposed basic VidHRFormer block. A temporal attention mask is applied to the Temporal MHSA module for VPTR-FAR. (b)VPTR-NAR. The left part is the Transformer encoder and right part is the non-autoregressive Transformer decoder.}
\end{figure*}

We proposed a new Transformer block based on the HRFormer block \cite{YuanFHLZCW21} for video processing, which is named VidHRFormer block. The detail architecture of a VidHRFormer block is shown in the gray area of Fig. \ref{fig:FAR_model}. Essentially, we integrate a temporal multi-head attention layer, together with some other necessary feed-forward and normalization layers, into the HRFormer block.

\textbf{Local spatial multi-head self-attention (MHSA).} Given a spatiotemporal feature map $Z \in \mathbb{R}^{N\times T\times H \times W \times d_{model}}$, we firstly reshape and evenly divide it into $P$ local patches $\{Z_1, Z_2, ..., Z_P\}$ along the $H$ and $W$ dimensions, where $Z_p \in \mathbb{R}^{(NT) \times K^2 \times d_{model}}$, and each local patch is of size $K\times K$,  with $P= \frac{HW}{K^2}$ patches in total. $MHSA(Z_p) = Concat[head(Z_p)_1, ..., head(Z_p)_h]$, where $head(Z_p)_i \in \mathbb{R}^{K^2\times \frac{d_{model}}{h}}$ is formulated as

\begin{equation}
    head(Z_p)_i = softmax[\frac{((Z_p^Q W_i^Q)(Z_p^K W_i^K)}{\sqrt{d_{model}/h}}]Z_p W_i^V,
\label{eq: attention}
\end{equation}

\noindent where $W_i^Q$, $W_i^K$, $W_i^V$ are linear projection matrices for the query, key and value  of each head $i$ respectively, $Z_p^Q$ and $Z_p^K$ denote the key and query for attention. We may use a fixed absolute 2D positional encoding \cite{carion2020}, or a relative positional encoding (RPE) \cite{shaw2018} of the local patch to get $Z_p^Q$ and $Z_p^K$. We compared the two different positional encodings in the experiments. The complexity of local spatial MHSA is $\mathcal{O}(\frac{H^2W^2}{P^2})$.

\textbf{Convolutional feed-forward neural network (Conv FFN).} After the local spatial MHSA, $\{Z_1, Z_2, ..., Z_P\}$ are assembled back to be $Z \in \mathbb{R}^{(NT)\times H \times W \times d_{model}}$. The Conv FFN layer is composed of a $3\times 3$ depth-wise convolution and two point-wise MLPs. Note that all the normalization layers in Conv FFN are layer normalization, instead of batch normalization used in the original HRFormer block.

\textbf{Temporal MHSA.} The local spatial MHSA and Conv-FFN are shared by every frame feature. A temporal MHSA is placed on top of them to model the temporal dependency between frames. We reshape the input feature map $Z \in \mathbb{R}^{(NT)\times H \times W \times d_{model}}$ to be $Z \in \mathbb{R}^{(NHW)\times T \times d_{model}}$. Temporal MHSA is a standard multi-head self-attention similar to the local spatial MHSA, except that there is no local patch division and it takes a fixed absolute 1D positional encoding of time. The complexity of temporal MHSA is $\mathcal{O}(T^2)$. The temporal MHSA is followed by a MLP feed-forward neural network as in a standard Transformer, and the output feature map is reshaped back to be $Z \in \mathbb{R}^{N\times T\times H \times W \times d_{model}}$ for the next layer of VidHRFormer block.

In summary, the proposed VidHRFormer block reduces the compute complexity from $\mathcal{O}((THW)^2)$ to be $\mathcal{O}(\frac{H^2W^2}{P^2} + T^2)$ by combining spatial local window attention and temporal attention in two steps. Based on the VidHRFormer, we develop two different VPTR models.

\subsection{VPTR-FAR}
\label{ssec: VPTR-FAR}
The fully autoregressive VPTR model is simply a stack of multiple VidHRFormer blocks. The architecture is shown in Fig. \ref{fig:FAR_model}. Theoretically, given a well-trained CNN encoder and decoder, the VPTR-FAR parameterizes the following distribution:
\begin{equation}
    p(x_1, ...,x_{L}, ..., x_{L+N}) = \prod_{t=1}^{L+N} p(x_t|x_{t-1}, ... x_1)
\end{equation}
In other words, VPTR-FAR predicts the next frame conditioned on all previous frames, which is also the most common paradigm for most SOTA VFFP models. An attention mask is applied to the temporal MHSA module to impose the conditional dependency between the next frame and previous frames.

During training, we feed the ground-truth frames $\{x_1, ..., x_{L+N-1}\}$ into the encoder, which generates the feature sequence $\{z_1, ..., z_{L+N-1}\}$. VPTR-FAR then predicts the future feature sequence $\{\hat{z}_2, ..., \hat{z}_{L+N}\}$, which is then decoded by the decoder to generate frames $\{\hat{x}_2, ..., \hat{x}_{L+N}\}$. The training loss of VPTR-FAR is:
\begin{equation}
    \mathcal{L}_{FAR} = \sum_{t=2}^{L+N}\mathcal{L}_2(x_t,\hat{x_t}) + \sum_{t=2}^{L+N} \mathcal{L}_{gdl}(x_t, \hat{x_t})
\label{eq: FAR_loss}
\end{equation}

 During test, we firstly get the ground-truth past frames features $\{z_1, ..., z_{L}\}$. Then there are two different ways of recurrently predicting the future frames. The first method is recurrently generating all the future frame features only by the VPTR-FAR module, i.e. $\hat{z}_{t} = \mathcal{T}(z_1, ..., z_{t-1}), t\in[L+1, ... L+N]$, where $\mathcal{T}$ denotes the VPTR-FAR module. Then we get $\hat{x}_t = Dec(\hat{z}_t), t\in[L+1, ... L+N]$, where $Dec$ denotes the CNN frame decoder. The second prediction method introduces two additional steps. Particularly, $\hat{z}_{t} = Enc(Dec(\mathcal{T}(z_1, ..., z_{t-1}))), t\in[L+1, ... L+N]$, where $Enc$ denotes the CNN frame encoder. In short, we decode each future feature to be frame $\hat{x}_t$, and then encode the frame back into a latent feature before the prediction of next future frame feature. The second way significantly reduces the accumulated error during inference, and the reasons are analyzed in the experiments section.

\subsection{VPTR-NAR}
Inspired by the achitecture of DETR \cite{carion2020}, a non-autoregressive variant is proposed to increase the inference speed and reduce inference accumulation error of autoregressive models. VPTR-NAR is illustrated in Fig. \ref{fig:NAR_model}. It consists of a Transformer encoder and decoder, where the encoder $\mathcal{T}_E$ encodes all past frame features $z_t, t\in[1, L]$ to be $e_{t}, t\in[1, L]$, which are normally named as "memories" in NLP. The architecture of $\mathcal{T}_E$, left part of Fig. \ref{fig:NAR_model}, is the same as the VPTR-FAR, except that there is no temporal attention mask for the temporal MHSA module. 

The decoder $\mathcal{T}_D$ of VPTR-NAR, right part of Fig. \ref{fig:NAR_model}, includes two more layers compared with $\mathcal{T}_E$. A temporal multi-head attention (MHA) layer and another output Conv FFN layer. The Temporal MHA layer is also called the encoder-decoder attention layer, which takes the memories as value and key, while the query is derived from the future frame query sequence $\{q_{L+1}, ..., q_{L+N}\}$, where $q_t\in \mathbb{R}^{H\times W\times C}, t\in [L+1, L+N]$. $\{q_{L+1}, ..., q_{L+N}\}$ is randomly initialized and updated during training. Note that there is no temporal attention mask for Temporal MHSA layer, since we do not need to impose conditional dependency between each future frame query. Theoretically, VPTR-NAR directly models the following conditional distribution:

\begin{equation}
    p(x_{L+N}, ..., x_{L+1}|x_{L}, ...,x_1)
\end{equation}

\textbf{Contrastive feature loss for VPTR-NAR.} We failed to train VPTR-NAR with a loss only composed by MSE and GDL, i.e., $\mathcal{L} = \sum_{t=L+1}^{L+N}\mathcal{L}_2(x_t,\hat{x_t}) + \mathcal{L}_{gdl}(x_t, \hat{x_t})$, since it is easy to fall into some local minimums. Specifically, all the predicted future frames are somewhat similar to each other. A similar phenomenon is also observed in the non-autoregressive NMT models, where the Transformer decoder frequently generate repeated tokens \cite{wang2019non}. To solve this problem, we impose another contrastive feature loss $\mathcal{L}_{c}$ \cite{andonian2021a} to maximize the mutual information between predicted future frame feature $\hat{z}_t$ and the future frame feature $z_t$ (ground-truth) generated by the CNN encoder, where $t\in [L+1, L+N]$. $\mathcal{L}_{c}$ is formulated as follows,

\begin{equation}
    \mathcal{L}_{c}(z_t, \hat{z}_t) = \frac{1}{2}\sum_{s=1}^{S_l}l_{c}(\hat{v}_s, v_s, sg(\bar{v}_s)) + l_{c}(v_s, \hat{v}_s, sg(\hat{\bar{v}}_s)),
\end{equation}

\noindent where $v_s \in \mathbb{R}^{d_{model}}$ denotes a feature vector at spatial location $s$ of $z_t$, $\bar{v}_s \in \mathbb{R}^{(S_l - 1)\times d_{model}}$ denotes the collection of feature vectors at all other spatial locations of $z_t$. $S_l = H\times W$ is the total number of spatial locations in a feature map. $\hat{v}_s$ and $\hat{\bar{v}}_s$ of $\hat{z}_t$ are defined in the same way. $sg$ is the stop gradient operation and $l_c$ is the info-NCE based contrastive loss defined by 

\begin{multline}
    l_{c}(v, v^+, v^-) = \\
    -log\frac{exp(s(v, v^+))}{exp(s(v, v^+)) + \sum_{m=1}^Mexp(s(v, v^-))}
\label{eq: NCE}.
\end{multline}
Given a feature vector $v \in \mathbb{R}^{d_{model}}$, $v^+ \in \mathbb{R}^{d_{model}}$ is the spatially-corresponding ground-truth feature vector, and $v^- \in \mathbb{R}^{M\times d_{model}}$ denotes the $M$ other spatially different ground-truth feature vectors. $s(v1, v2)$ measures the feature dot-product similarity. Finally, the training loss function for VPTR-NAR is defined as 

\begin{equation}
    \mathcal{L}_{NAR} = \sum_{t=L+1}^{L+N}\mathcal{L}_2(x_t,\hat{x_t}) + \mathcal{L}_{gdl}(x_t, \hat{x_t}) + \lambda_2 \mathcal{L}_{c}(z_t, \hat{z}_t).
\label{eq: NAR_loss}
\end{equation}

During test, VPTR-NAR predicts $N$ future frames simultaneously, instead of recurrently.

\subsection{Training strategy} 
The whole VFFP model training process is divided into two stages. For stage one, we ignore the VPTR module and only train the encoder and decoder as a normal autoencoder with the loss function in Eq. \ref{eq: AE_loss}, which aims to reconstruct all the frames of the whole training set perfectly. During stage two, we only update parameters of the VPTR module while the well-trained encoder and decoder are fixed. VPTR-FAR and VPTR-NAR are trained with the loss function in Eq. \ref{eq: FAR_loss} and Eq. \ref{eq: NAR_loss} respectively. It is well-known that Transformers are hard to train, therefore we proposed this two-stage training strategy to ease the training. Besides, the two-stage training strategy is flexible and allows us to test different VPTR variants without repetitive training of the encoder and decoder. Experimental results show that a final joint finetuning of autoencoder and VPTR is not helpful.

\section{Experiments}
\subsection{Datasets and Metrics}
We evaluate the proposed VPTR models over three datasets, KTH \cite{schuldt2004}, MovingMNIST \cite{srivastava2015} and BAIR \cite{ebert2017}. For KTH and Moving MNIST, VPTR models are trained to predict 10 future frames given 10 past frames. For BAIR dataset, VPTR models are trained to predict 10 future frames given 2 past frames. All datasets are trained with a resolution of $64\times 64$. We process the KTH dataset as previous works \cite{villegas2017a, jin2020}. Random horizontal and vertical flip of each video clip are utilized as data augmentation. We use the MovingMNIST created by E3D-LSTM \cite{wang2018a}, which takes the same data augmentation method as KTH. There is no data augmentation for BAIR.

\textbf{Metrics.} Learned Perceptual Image Patch Similarity (LPIPS)\cite{zhang2018} and Structural Similarity Index Measure (SSIM) are used to evaluate all the three datasets. Peak Signal-to-Noise Ratio (PSNR) is used to evaluate the KTH and BAIR dataset, and Mean Square Error (MSE) is used to evaluate the MovingMNIST dataset. All the LPIPS values are presented in $10^{-3}$ scale.

\subsection{Implementation}
\textbf{Training stage one.} In Eq. \ref{eq: AE_loss}, $\lambda_1 = 0.01$ for KTH and MovingMNIST, $\lambda_1 = 0$ for the BAIR dataset. The optimizer is Adam \cite{Kingma2015}, with a learning rate of $2e^{-4}$. \textbf{Training stage two.} For the visual features of each frame, $H=8, W=8, d_{model}=528$. $K = 4$ for the local spatial MHSA. The Transformer of VPTR-FAR includes 12 layers. For VPTR-NAR, the number of layers of $\mathcal{T}_E$ is 4, and the number of layers of $\mathcal{T}_D$ is 8. We take AdamW \cite{loshchilov2018} with a learning rate of $1e^{-4}$ for the optimization of all Transformers. Gradient clipping is taken to stabilize the training. For the loss function of VPTR-NAR (Eq. \ref{eq: NAR_loss}), $\lambda_2 = 0.1$.

\subsection{Results}

\begin{figure}[h]
\centering
\includegraphics[clip, trim=5.4cm 6.6cm 5cm 3.5cm, width=\linewidth]{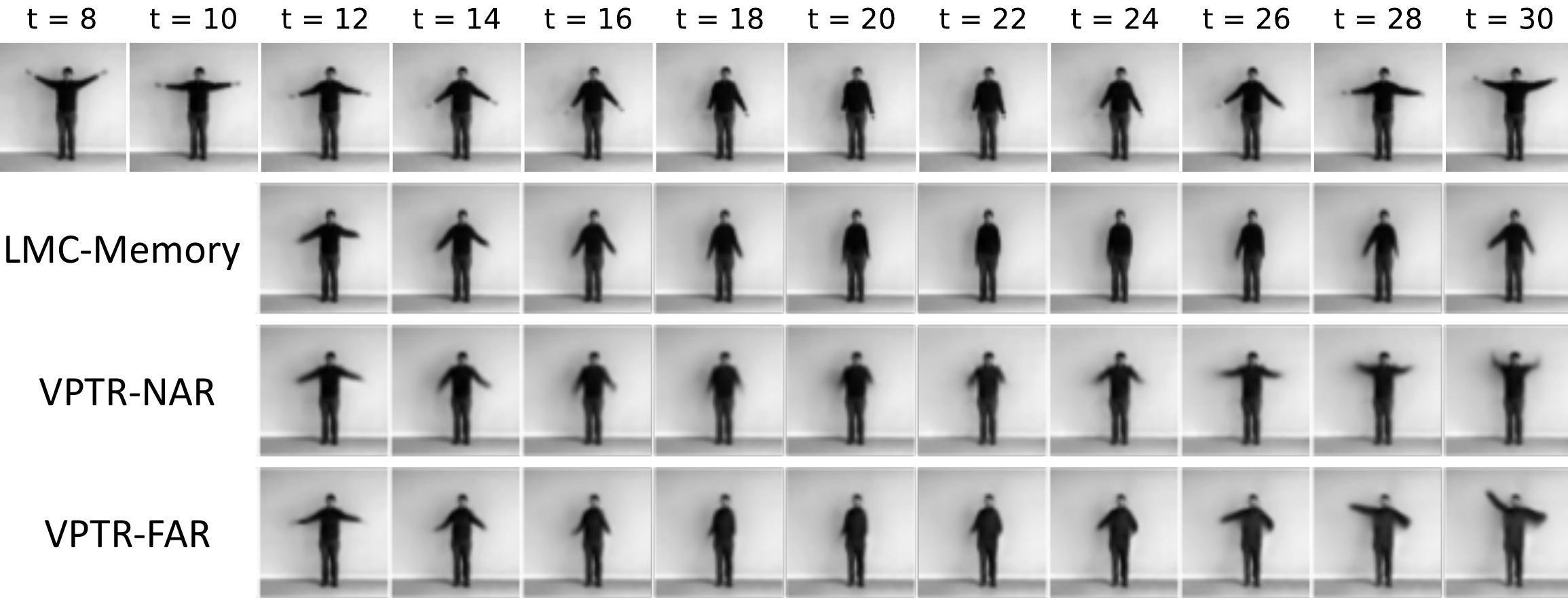}
\caption{Qualitative results on KTH dataset. The first row is ground-truth. For the past frames, $t\in [1, 10]$. For the future frames, $t\in [11, 30]$. }
\label{fig:KTH_results}
\end{figure}

\textbf{Results on KTH.} The best results of the two VPTR variants are recorded in Table \ref{tab:kth-movingmnist}. Following the evaluation protocol of previous works, we extend the prediction length to be 20 frames during test. Compared with the SOTA models, the proposed VPTR models reach competitive performances in terms of PSNR and SSIM. Notably, both two VPTR variants outperform the SOTAs in terms of LPIPS by a large margin. Some prediction examples are shown in Fig. \ref{fig:KTH_results}. It shows that the predicted arm motion by VPTRs is more aligned with the ground-truth, which indicates that the VPTRs more successfully capture the cyclic hand waving movements that only depends on the past frames, in contrast to LMC-Memory that recalls some inaccurate motion from the memory bank. 

\begin{table}[h]
\caption{Results on KTH and MovingMNIST. $\uparrow$: higher is better, $\downarrow$: lower is better. \textbf{Boldface}: best results.}
\centering
\begin{tabular}{cccccc|ccc} \hline
\multicolumn{3}{c}{\multirow{3}{*}{Methods}} & \multicolumn{3}{c}{KTH} & \multicolumn{3}{c}{MovingMNIST} \\
& & & \multicolumn{3}{c}{$10 \rightarrow20$} & \multicolumn{3}{c}{$10 \rightarrow 10$} \\
& & & PSNR$\uparrow$ & SSIM$\uparrow$ & LPIPS$\downarrow$ & MSE$\downarrow$ & SSIM$\uparrow$ & LPIPS$\downarrow$ \\ \hline
\multicolumn{3}{c}{MCNET \cite{villegas2017a}}& 25.95 & 0.804 & - & - & - & - \\
\multicolumn{3}{c}{PredRNN++ \cite{wang2018d}} & 28.47 & 0.865 & 228.9 & 46.5 & 0.898 & 59.5 \\
\multicolumn{3}{c}{E3D-LSTM \cite{wang2018a}} & 29.31 & 0.879 & - & \textbf{41.3} & 0.910 & - \\
\multicolumn{3}{c}{STMFANet \cite{jin2020}} & \textbf{29.85} & 0.893 & 118.1 & - & - & - \\
\multicolumn{3}{c}{Conv-TT-LSTM \cite{su2020a}} & 28.36 & \textbf{0.907} & 133.4 & 53.0 & 0.915 & \textbf{40.5} \\
\multicolumn{3}{c}{LMC-Memory \cite{lee2021}} & 28.61 & 0.894 & 133.3 & 41.5 & \textbf{0.924} & 46.9 \\ \hline
\multicolumn{3}{c}{VPTR-NAR} & 26.96 & 0.879 & 86.1 & 63.6 & 0.882 & 107.5 \\
\multicolumn{3}{c}{VPTR-FAR} & 26.13 & 0.859 & \textbf{79.6} & 107.2 & 0.844 & 157.8 \\ \hline

\end{tabular}
\label{tab:kth-movingmnist}
\end{table}

\textbf{Results on MovingMNIST.} The right part of Table \ref{tab:kth-movingmnist} shows the results on the MovingMNIST dataset. We observe that the SSIM of the VPTR variants is close to the SOTAs, but there are large gaps in terms of MSE and LPIPS, especially for VPTR-FAR.
Qualitative examination shows that VPTRs make poor predictions for the overlapping characters.

\textbf{Results on BAIR.} Compared with KTH and MovingMNIST, BAIR is more challenging, because the robot arm motion is random and only two past frames are given for the prediction. From Table \ref{tab:bair}, we find that VPTR-NAR outperforms STMFANet \cite{jin2020} in terms of SSIM and LPIPS. Our good performance can be attribute to the large model capacity of VPTRs and the large size of BAIR dataset. We note however that the predicted robot arm becomes blurry after the first few frames, due the deterministic nature of VPTRs. Our VPTRs could be extended to be stochastic models easily, and we expect that the stochastic version of VPTRs would achieve an even better performance on the BAIR dataset.

\begin{table}[h]
\caption{Results on BAIR. $\uparrow$: higher is better, $\downarrow$: lower is better. \textbf{Boldface}: best results.}
\centering
\begin{tabular}{cccccc} \hline
\multicolumn{3}{c}{Methods} & PSNR$\uparrow$ & SSIM$\uparrow$ & LPIPS$\downarrow$ \\ \hline
\multicolumn{3}{c}{SV2P \cite{babaeizadeh2018}} & 20.36 & 0.817 & 91.4 \\
\multicolumn{3}{c}{SVG-LP \cite{denton2018}} & 17.72 & 0.815 & 60.3 \\
\multicolumn{3}{c}{Improved VRNN \cite{castrejon2019}} & - & 0.822 & 55.0 \\
\multicolumn{3}{c}{STMFANet \cite{jin2020}} & \textbf{21.02} & 0.844 & 93.6 \\ \hline
\multicolumn{3}{c}{VPTR-NAR} & 19.40 & \textbf{0.852} & \textbf{53.9} \\
\multicolumn{3}{c}{VPTR-FAR} & 18.63 & 0.824 & 69.3 \\ \hline
\end{tabular}
\label{tab:bair}
\end{table}

\subsection{Ablation Study}

\begin{table}[h]
\caption{Ablation study on KTH and MovingMNIST. $\uparrow$: higher is better, $\downarrow$: lower is better. \textbf{Boldface}: best results.}
\centering
\begin{tabular}{ccccccccc} \hline
\multicolumn{3}{c}{\multirow{3}{*}{Methods}} & \multicolumn{3}{c}{KTH} & \multicolumn{3}{c}{MovingMNIST} \\
& & & \multicolumn{3}{c}{$10 \rightarrow20$} & \multicolumn{3}{c}{$10 \rightarrow 10$} \\
& & & PSNR$\uparrow$ & SSIM$\uparrow$ & LPIPS$\downarrow$ & MSE$\downarrow$ & SSIM$\uparrow$ & LPIPS$\downarrow$ \\ \hline

\multicolumn{3}{c}{VPTR-NAR-BASE} & 26.92 & \textbf{0.881} & 94.6 & 64.2 & 0.880 & 114.2 \\
\multicolumn{3}{c}{VPTR-NAR-RPE} & \textbf{26.96} & 0.879 & 86.1 & \textbf{63.6} & \textbf{0.882} & \textbf{107.5} \\
\multicolumn{3}{c}{VPTR-NAR-FEDA} & 26.25 & 0.872 & 101.1 & 68.0 & 0.872 & 128.7 \\\hline
\multicolumn{3}{c}{VPTR-FAR-BASE} & 25.71 & 0.816 & \textbf{79.5} & 108.3 & 0.843 & 157.3 \\ 
\multicolumn{3}{c}{VPTR-FAR-RPE} & 26.13 & 0.859 & 79.6 & 107.2 & 0.844 & 157.8 \\ 
\multicolumn{3}{c}{VPTR-FAR-RIL} & 21.61 & 0.678 & 192.7 & 138.2 & 0.821 & 445.7 \\ \hline

\end{tabular}
\label{tab:ablation}
\end{table}

\textbf{RPE.} The VPTRs with fixed absolute positional encodings are taken as the base models, i.e. VPTR-NAR-BASE and VPTR-FAR-BASE in Table \ref{tab:ablation}. To investigate the influence of relative positional encodings, we get VPTR-NAR-RPE and VPTR-FAR-RPE by substituting the 2D absolute positional encoding of all local spatial MHSA module with a learned 2D RPE. We argue that RPE is beneficial because both VPTR-FAR-RPE and VPTR-NAR-RPE outperform the base models with regard to most metrics on the two datasets.

\textbf{Spatial-temporal separation attention.}
The separation of spatial and temporal attention reduces the complexity, but it also means that a feature at one location only attends to partial locations of the whole spatiotemporal space. To investigate the influence of the separated attention, we replace the encoder-decoder attention layers of VPTR-NAR with a full spatiotemporal attention, which has a complexity of $\mathcal{O}(\frac{H^2W^2T^2}{P^2})$. The increased computation cost is affordable as we only replace the encoder-decoder attention layers. Comparing the VPTR-NAR-FEDA with the base model, where FEDA denotes ``full encoder-decoder attention", we find that FEDA is not beneficial. It indicates that the alternate stacking of multiple spatial and temporal attention layers is capable of propagating global information from past frames to future frames.

\textbf{Autoregressive inference methods.} As we have described in Section \ref{ssec: VPTR-FAR}, we can perform recurrently inference over latent space (RIL) or recurrently inference over pixel space (RIP) for VPTR-FAR. VPTR-FAR-BASE is evaluated by RIP. Even though RIL is little faster than RIP, VPTR-FAR-BASE outperforms VPTR-FAR-RIL by a large margin. Severe accumulation of errors is observed for VPTR-FAR-RIL.

We believe the reason is that VPTR-FAR receive only supervision from the pixel space during training. There is no direct constraints on the distance between the feature space predicted by the Transformer and the feature space generated by the CNN encoder. Furthermore, the latent space dimension of the autoencoder is greater than the pixel space dimension, which is a common case for VFFP, as we expect a good reconstruction visual quality. Therefore, recurrent inference only depending on the Transformer predictor would make the predicted features deviate from the ground-truth (learned by autoencoder during stage one) features quickly. But decoding the feature firstly and then encoding it back into latent space restrict the deviation to some degree.

\subsection{Comparison of VPTR variants}

\begin{figure}[h]
\centering
\includegraphics[clip, trim=0.3cm 0.2cm 0.2cm 0.2cm, width=0.8\linewidth]{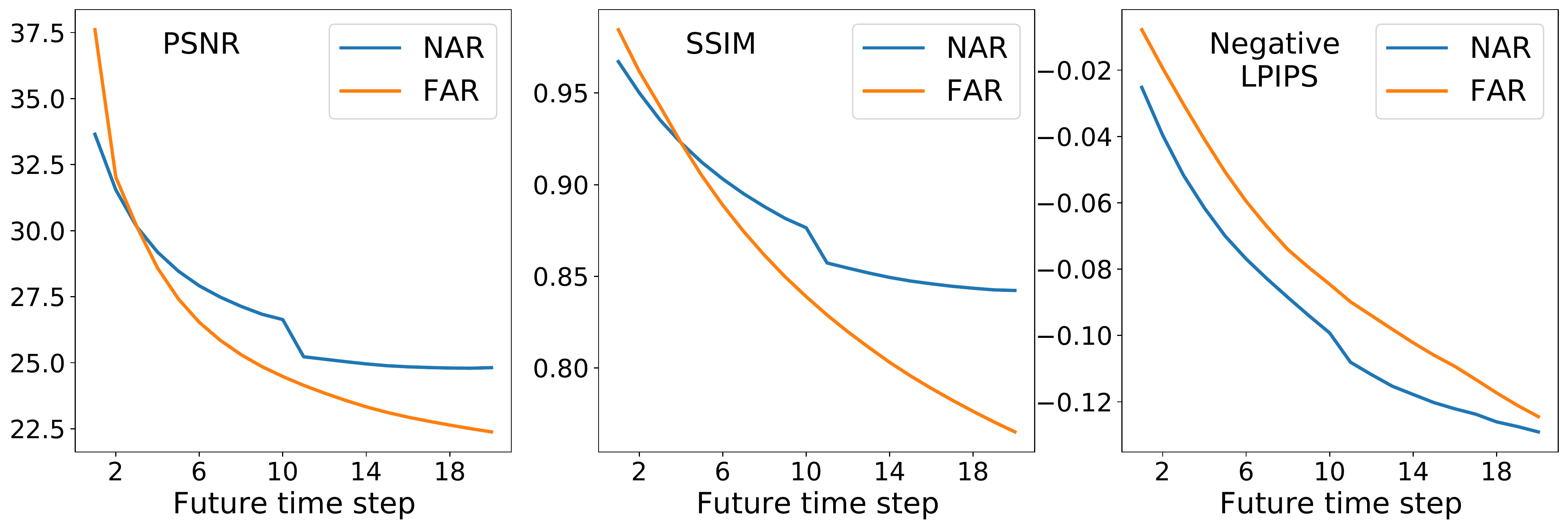}
\caption{Results of VPTR variants on KTH for increasing prediction steps.}
\label{fig:variants_comparison}
\end{figure}

For a better visualization, we plotted the metrics curve of VPTR-NAR-BASE and VPTR-FAR-BASE with respect to the predicted future time steps in Fig. \ref{fig:variants_comparison}. For the first few predicted frames, VPTR-FAR achieves a better PSNR and SSIM than VPTR-NAR, but the values drop quickly due to the accumulated errors introduced by recurrent inference. It shows that VPTR-NAR has a smaller quality degradation in terms of PSNR and SSIM. For the last 10 steps of the LPIPS curve, VPTR-NAR also has a smaller slope than VPTR-FAR. 

The accumulation of errors in VPTR-FAR is mainly due to the discrepancy between training and testing behaviors. Specifically, the previously predicted frames are used during inference instead of the ground-truth as during training, which leads to a worse generalization ability of VPTR-FAR given the same number of Transformer layers as VPTR-NAR. In contrast, there is no discrepancy the between training and testing behaviors of VPTR-NAR. However, it is more difficult for the VPTR-NAR to estimate the joint distribution directly, so an additional contrastive feature loss is required.

Another advantage of VPTR-NAR is the faster inference speed. For VPTR-NAR, predicting $N$ frames has a complexity of $\mathcal{O}(N^2)$, but the complexity for VPTR-FAR is $\mathcal{O}(\sum_{n=1}^N n^2)$. For simplicity, in this assessment, we ignored the spatial dimensions of features, computation cost of processing past frames, and supposed that the future frames length of inference is same as of the training. However, the model size of VPTR-NAR is larger because of the learned future frame queries.

\section{Conclusion}
In this paper, we proposed an efficient VidHRFormer block for spatiotemporal representation learning, and two different VFFP models are developed based on it. We expect that the proposed VidHRFormer block could be used as a backbone for many other video processing tasks. We compared the performance of proposed VPTRs with SOTA models on various datasets, and we are competitive with more complex models. Finally, we analyzed the influence of different modules for two VPTR variants by a thorough ablation study, and we observed that VPTR-NAR achieves a better performance than VPTR-FAR.






%

\bibliographystyle{IEEEtran}
\bibliography{references}



\end{document}